\documentclass[11pt,a4paper]{article}
\usepackage[hyperref]{emnlp-ijcnlp-2019}
\usepackage{times}
\usepackage{latexsym}

\usepackage{url}

\usepackage{amsmath}
\usepackage{algorithm}
\usepackage{algorithmic}
\usepackage{url}
\usepackage{adjustbox}
\usepackage{multirow}
\usepackage{tabularx}
\usepackage{hhline}

\aclfinalcopy 

\title{Identifying and Explaining Discriminative Attributes}

\author{Armins Stepanjans \and Andr\'e Freitas\\
       Department of Computer Science \\
       University of Manchester\\
       \texttt{armins.stepanjans@student.manchester.ac.uk} \\
       \texttt{andre.freitas@manchester.ac.uk}
}

\date{}

\begin{document}
\maketitle
\begin{abstract}
Identifying what is at the center of the meaning of a word and what discriminates it from other words is a fundamental natural language inference task. This paper describes an explicit word vector representation model (WVM) to support the identification of discriminative attributes. A core contribution of the paper is a quantitative and qualitative comparative analysis of different types of data sources and Knowledge Bases in the construction of explainable and explicit WVMs: (i) \textit{knowledge graphs built from dictionary definitions}, (ii) \textit{entity-attribute-relationships graphs derived from images} and (iii) \textit{commonsense knowledge graphs}. Using a detailed quantitative and qualitative analysis, we demonstrate that these data sources have complementary semantic aspects, supporting the creation of explicit semantic vector spaces. The explicit vector spaces are evaluated using the task of discriminative attribute identification, showing comparable performance to the state-of-the-art systems in the task (F1-score = 0.69), while delivering full model transparency and explainability.
\end{abstract}

\section{Introduction}

Word-vector/embedding models (WVM) have emerged as first-class representations in contemporary Natural Language Processing (NLP) tasks due to their ability to capture semantic similarity and relatedness in an unsupervised and comprehensive manner \cite{turney2010frequency}, \cite{freitas2015schema}. Additionally, the simplicity entailed by the vector space abstraction makes it an engineering-friendly representation, also explaining its widespread adoption and use \cite{freitas2015schema}. However, the latent features (dense vectors) at the center of most of the best-performing models have limited their application to two main uses: (i) computing semantic similarity and relatedness measures and (ii) performing vocabulary generalization as an input layer on Machine Learning (ML) models.

The identification of discriminative attributes (IDA), recently introduced by \citet{semeval2018task10}, can motivate the development of word vector models which can support types of operations with finer semantics, going beyond the computation of semantic similarity and relatedness scores, with potential applications in fine-grained semantic inference tasks.

Concretely, using the example provided by \cite{semeval2018task10}, given a pair of target terms \emph{apple} and \emph{banana}, the IDA task seeks to answer if the term \emph{red} is a discriminative attribute for apple in comparison to banana. According to Krebs et al. \emph{semantic difference} is a ternary relation between two concepts (\textit{apple}, \textit{banana}) and a discriminative feature (\textit{red}) that characterizes the former concept but not the latter.

This paper focuses on proposing an explicit (sparse) WVM for detecting and explaining discriminative attributes. The proposed explicit WVM provides a dimension of \textit{explainability} while keeping the simplicity of the vector space representation model. This addition allows the model to provide a \textit{justification} while addressing tasks such as the computation of discriminative attributes (Figure \ref{indxDiagram}). The explicit nature of the model and its ability to compute the difference between the term pairs is the core proposed contribution (instead of the improvement of the F1-score for the task). Models with the ability to compute discriminative attributes can provide a representation paradigm which can support more fine-grained semantic inference tasks.  

Another core contribution of this paper is a comprehensive quantitative and qualitative comparative analysis on how different types of data sources and Knowledge Bases (KBs) affect the construction of explainable and explicit WVMs. The IDA task requires a granular representation of the necessary and sufficient conditions associated with the definition of a target concept, many of these transcending lexicographic and encyclopaedic representations. In order to address this problem, we analyse three types of data sources: (i) \textit{knowledge graphs built from dictionary definitions}, (ii) \textit{entity-attribute-relationship graphs derived from images} and (iii) \textit{commonsense knowledge graphs}. The goal behind using these models is twofold: (i) to provide an explainable, fine-grained word-vector model and (ii) to support the capture of lexical-semantic relations not captured in a representative fashion in regular corpora used in existing distributional models (which in most cases combine news, encyclopedic and literary discourse). 

In summary, this paper focuses on the following contributions: \textbf{(i)} \textit{The creation of a novel interpretable word vector model based on the combination of structured definitions and visual features for the IDA task} and \textbf{(ii)} \textit{a detailed comparative, quantitative and qualitative evaluation of the semantic contribution of different types of data sources for explicit semantics}.

\section{Related Work}
\label{sec:relatedWork}

In this section we group related work into two major categories: (i) approaches for the identification of discriminative attributes  (IDA) and (ii) definition-based word vector space models.

Existing approaches have explored combinations of linguistic and data resources (WordNet, ConceptNet, Wikipedia), linguistic features (syntactic dependencies), sparse word vector models (JoBimText), dense word-vector (DWV) models (W2V, GloVe) and supervised/ unsupervised machine learning approaches (SVM, MLP, Ensemble Methods).

With regard to interpretability and explainability we can classify IDA approaches into three categories. \textit{Frequency-based models over text-based features}, heavily relying on textual features and frequency-based methods \cite{citiusnlp, elirfupv} ; \textit{ML over Textual features} \cite{alb, meaningspace, unbnlp, abdn} and \textit{ML over dense vectors and textual features} \cite{uwb, ghh, thungn, alb,unam, luminoso, bomji, igevorse, ecnu, amritanlp, discriminator, umd, ntunlp}. While the first category concentrates on models with higher interpretability, none of these models provide explanations.

In the area of definition-based word vectors, similar initiatives were concentrated in the areas of definition-based distributional models and interpretable distributional semantic models.  \citet{baroni2010strudel} describes an approach that automatically constructs a distributional semantic model from a text corpus. The model represents concepts in terms of weighted typed properties and is capable of describing the similarities between concepts as well as the properties responsible for this similarity. \citet{murphy2012learning} apply a matrix-factorization technique (NNSE) to produce sparse embeddings. In addition, their embeddings are interpretable in a way, that given a dimension in the vector space, vectors in that dimension have a non-latent relatedness to a human-interpretable concept. Their work was extended to be able to form composed representation of word phrases while still maintaining the desired interpretability of the original model \cite{fyshe2015compositional}. %

Comparatively, this work focuses on the creation of an explicit word vector space model (EWVM), with an associated explanation, evaluating the performance of different types of lexico-semantic resources in the context of the task of identification of discriminative attributes (IDA).

\section{Identifying and Explaining Discriminative Attributes}

\subsection{Problem Definition}

This paper provides an explainable word vector space model (EWVM) for detecting whether a given term is a discriminative attribute or not with regard to a pair of reference terms as defined by \citet{semeval2018task10}. Given a triple $<t_p, t_c, t_f>$,  $t_f$ (\emph{feature}) is considered discriminative if it is related to the first term $t_p$ (\emph{pivot}) to a significantly higher extent than it is related to the second term $t_c$ (\emph{comparison}), i.e. \mbox{$t_f \in p(t_p) \land t_f \notin p(t_c)$}, where the function $p(t)$ returns a set of properties associated with term $t$. 

Additionally, beyond the identification of the discriminative attribute (i.e. assigning a value of true or false to the target term triple), this paper explores different notions of an explanation $e$ assigned to the inference, which can be both human and machine interpretable.

\subsection{Types of Discriminative Attributes}

The nature of the semantic relationships expressed in the task of identifying discriminative attributes can guide the selection of the supporting target corpora. In this section, we propose a classification of \textit{term-discriminative attribute relationships} into three dual categories: \textbf{Essential vs. Incidental}, \textbf{Sensory vs. Logical} and \textbf{Relative vs. Absolute}. 

\paragraph{Essential vs. Incidental:} $p$ is an \emph{essential} property of an object $o$ just in case it is necessary that $o$ has $p$, whereas $p$ is an \emph{incidental} property of an object $o$ just in case $o$ has $p$ but it is possible that $o$ lacks $p$. Example: \texttt{(cognac, whiskey, french)} is \emph{essential} and \texttt{(nose, throat, perfume)} is \emph{incidental}. The notion of Essential v. Incidental is similar to the logical notion of Necessary and Sufficient Conditions, but rooted in the philosophical notion of essential and accidental properties, particularly, as defined by \citet{robertson:1}.

\paragraph{Sensory vs. Logical:} $p$ is a \emph{sensory} property of an object $o$, in case $p$ can be identified as a property of $o$ exclusively through sensory information, whereas $p$ is a \emph{logical} property of an object $o$ otherwise. Example: \texttt{(cheek, brow, red)} is \emph{sensory} and \texttt{(grapes, wine, fruit)} is \emph{logical}. The Sensory vs. Logical dual category is inspired by Hayes' Second Na{\"i}ve Physics Manifesto, particularly, the distinction of `three ways in which tokens can be attached to their denotations', where our notion of Sensory attribute corresponds to the second way (`some of the tokens can be attached to sensory and motor systems'), whereas our notion of Logical attribute corresponds to the remaining two (`tokens could be attached to the world through language' and `token is a metatheory of some internal part of the theory') \cite{hayes:1}.

\paragraph{Relative vs. Absolute:} $p$ is a \emph{relative} property of an object $o_{a}$ in relation to $o_{b}$ just in case there exists an object $o_{c}$ with the property $p$ where $p$ is not an attribute of $o_{a}$ in relation to $o_{c}$. Whereas $p$ is an \emph{absolute} property of an object $o_{a}$ in case there does not exist an object $o_{b}$ with the property $p$ where $p$ is not an attribute of $o_{a}$ when compared to $o_{b}$. Example: \texttt{(giraffe, ostrich, tall)} is \emph{relative} and \texttt{(bat, butterfly, fur)} is \emph{absolute}. The Relative vs. Absolute dual category follows Hayes' idea of `intrinsic qualities (absolute) versus the distance between such qualities (relative)' \cite{hayes:1}.

The classification scheme will guide the creation of the discriminative word vector model described in the next section.

\section{Building Explicit Word Vector Space Models (EWVM)}

Based on the dual categories identified in the previous section, a composition of three word vector space models is used to define \emph{explicit word vector spaces} (EWVM), using three different types of corpora:

\noindent \textbf{Definition-Based Model (DBM):} Consists of a dictionary-style definition corpus. In the context of this work, WordNet and Wiktionary \textit{natural language definitions} are used as data sources.

\noindent \textbf{Visual Feature Model (VFM):} Built using lexical graph descriptors for images, containing entities, attributes and relations, grounded on image bounding boxes.

\noindent \textbf{Commonsense Knowledge Graph (CKG):} Consists on the use of lexico-semantic knowledge graphs such as ConceptNet.

\noindent The following sections describe the construction of the EWVM.

\subsection{Definition-Based Models (DBMs)}

DBMs are built out of natural language term definitions (glosses) found either in dictionaries or filtered out of larger corpora (for example, Wikipedia contains many definitional sentences which can be isolated using lexico-syntactic patterns). The intuition behind the use of natural language definitions is twofold: first they are succinct descriptions of the necessary attributes associated to a concept and secondly they are abundant across domains and languages as dictionaries or definitions embedded within discourse. The latter point makes DBMs potentially transportable across languages and domains.

\subsubsection{Model Construction}

\begin{figure*}[t]
	\centering
    \includegraphics[width=.85\textwidth]{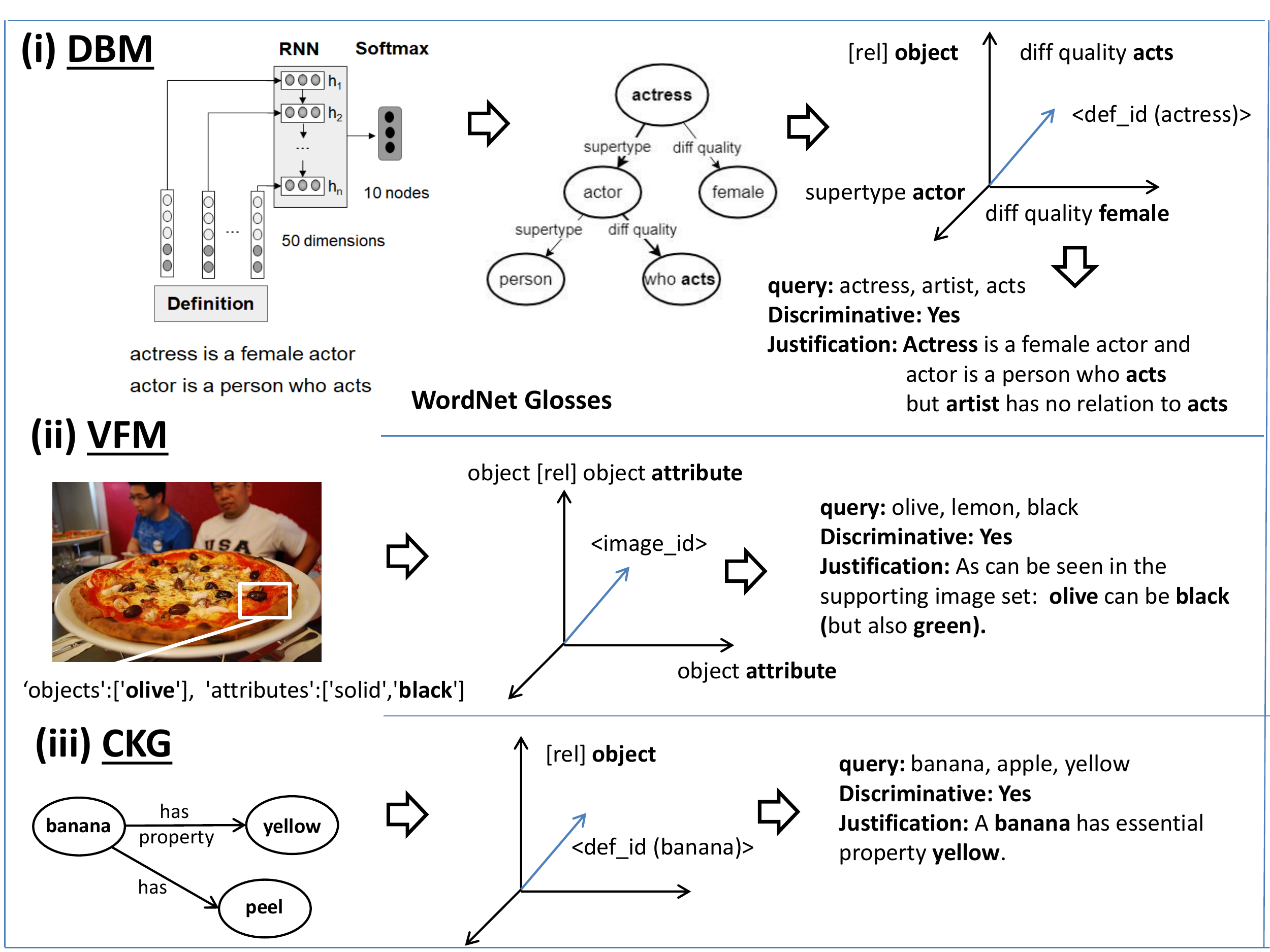}
    \caption{Sparse vector space representation and explanation of each model component.}
    \label{indxDiagram}
\end{figure*}

The representation behind the definition-based model is built by segmenting and categorizing natural language definitions into a set of semantic roles, a model which was introduced by \shortcite{silva:1}, \shortcite{silva:2}, \shortcite{VivianKG}. These roles aim to transform natural definitions into definition knowledge graphs, in order to facilitate natural language inference tasks \cite{silva:2}, \cite{VivianComp}. The set of semantic roles includes: \emph{supertype, differentia quality, differentia event, event location, purpose, accessory determiner, origin location}. Figure \ref{indxDiagram} (DBM) depicts an example of a classified definition. 

The semantic roles are assigned by building a recurrent neural network (RNN) Definition Role Labeling (DRL) classifier using POS-tags, and pre-trained word vectors as features using the configuration of Silva et al. \shortcite{silva:1}. After the semantic roles are assigned, they are used as an input to build the supporting word vector space. For each definition segment, all tokens are lemmatized and stop-words removed. Afterwards, an inverse document frequency (idf) weighting scheme is applied, in order to support the computation of semantic similarity and relatedness scores within the model (despite not being the target use of the model). Additionally, for each target term, we take into account its upward taxonomic chain, i.e. it inherits the definition attributes from the parent terms linked by the detected supertypes at the definition. %

An inverted index is used to materialize the vector space. A workflow of the proposed model is depicted in Figure \ref{indxDiagram}.

\begin{algorithm}
 	\caption{EDAM: Identifying discriminative attributes}
     \label{alg:comb}
    \begin{algorithmic}
    	\STATE{$D$: Set of definitions $d$, containing predicates $p^D$ and terms $t$}
        \STATE{$I$: Set of images $i$, containing features with predicates $p^{[O,R,A]}$ and terms $t$}
		\STATE{\textbf{Query:} $<t_p, t_c, t_a>$}
        \STATE{\textbf{Output:} discriminative, explanation}
          \IF{$ (t_a, t_p) \in D \land (t_a, t_c) \notin D$}
            \STATE{discriminative $\leftarrow$ \textbf{true}}
            \STATE{explanation $\leftarrow$ template($t_p, t_c, t_a, d$)}
            \RETURN
          \ENDIF
        
          \IF{$ (t_a, t_p) \in I \land (t_a, t_c) \notin I$}
            \STATE{discriminative $\leftarrow$ \textbf{true}}
            \STATE{explanation $\leftarrow$ template($t_p, t_c, t_a, i$)}
            \RETURN
          \ENDIF

		  \STATE{discriminative $\leftarrow$ \textbf{false}}
    \end{algorithmic}
\end{algorithm}

\subsection{Visual Feature Model}

Natural language definition corpora by design focus on \textit{essential}, \textit{logical} and \textit{absolute} discriminative attribute types. However, discriminative attributes can also occur as \textit{incidental}, \textit{sensorial} or \textit{relative} instances. Most distributional semantic models available today are built over journalistic, encyclopedic or narrative types of discourse. While incidental attributes can be captured as a second-order distributional phenomena, these corpora do not reflect explicit commonsense knowledge, in particular with regard to extra-linguistic phenomena. 

Recent datasets introduced for the purpose of supporting image classification tasks, such as VisualGenome \cite{krishna2017visual} have provided multi-modal resources connecting sub-symbolic visual data types to symbolic-level categories. These datasets explore both modalities of visual and textual data. 

VisualGenome is a dataset consisting of scenes (108,077 images) segmented into bounding boxes (5.4M) and annotated with a set of objects (3.8M), attributes (2.8M) and relationships (2.3M). Each image has an associated lexical-semantic model represented as a labeled graph. VisualGenome concentrates on the description of a large spectrum of commonsense scenes (not focusing on specific named entities). Common terms are: \emph{man, person, tree, window, grass, table} (objects), \emph{white, black, blueish, metallic, round} (attributes) and \emph{along, inside, almost, above, ride} (relationships). Essentially, VisualGenome expresses facts about objects in commonsense scenes.

The commonsense nature of VisualGenome scenes and the \emph{object-attribute} relations provide a foundation for covering visual attribute sets, potentially covering part of the \emph{sensorial} attributes. Additionally, VisualGenome can be used to cover \emph{incidental} attributes. VisualGenome can provide some level of support for identifying relative discriminative attributes, by focusing on features which are mediated by visual interpretation such as size (e.g. large, small). However, this work does not focus on a representation which can support the identification of relative attributes mediated by visual features.

\subsubsection{Model Construction}

The Visual Feature Model (VFM) also uses a sparse explicit semantic vector space representation as its basis. The representation targets the identification of sensorial and incidental discriminative attributes, with a supporting explanatory model. 

The model is based on the construction of two vector spaces: one for indexing objects and attributes, \emph{object-attribute} (OA) space associated with bounding boxes and the other for indexing relationships, i.e. the \emph{scene-object-relationship} (SOR) space. The OA space supports the identification of the set of attributes directly associated with objects while the SOR space supports the capture of association relations between objects (\textit{objectA} can inherit an attribute from \textit{objectB}).

\subsection{Commonsense Knowledge Graph}

Dictionary definitions have a limited ability to express incidental relations among concepts due to their conciseness. For example, relations which express affordances (e.g. `can be used for') are not typically expressed in dictionaries. Labeled visual datasets are limited in their domain coverage. We use commonsense knowledge graphs as a third data source aiming to fill this gap. For example Visual Genome contains 5.4 million region descriptions, whereas ConceptNet 5 contains 28 million relationships, captured using 39 relationship types.

\subsubsection{Model Construction}

The Commonsense Knowledge Graph is mapped into a sparse explicit semantic vector space. The model targets essential and logical attributes. The model is constructed by indexing each relationship using an idf weighting scheme, while ignoring negation (prefixed by “Not”).

\subsection{Combined Model: Explainable Discriminative Attribute Model (EDAM)}

The goal of the EDAM model is twofold: to maximize the quality of the discriminative attribute identification (IDA) and to maximize the underlying interpretability of the model. The EDAM model combines all three models by composing them into an interpretability hierarchy.

Each component of the EDAM model provides a different type of explanation, described below:

\noindent \textbf{DBM/CKG}: Consists of an attribute path between the natural language definitions, using the structure defined by the  graphs. It provides an explanatory model at the intensional level.

\noindent \textbf{VFM}: Consists of the description of the attribute-pairs incidence and the supporting ground image-set. It provides an explanatory model at an extensional level.

The explanations are generated using a template-based approach as described in Algorithm \ref{alg:comb}. Details on the construction of the models and on the formalization of the explanations can be found online\footnote{\url{https://github.com/ab-10/Hawk}}.

\section{Evaluation}
\label{sec:Eval}

The proposed EDAM model was evaluated using the Semeval2018 Task 10 gold-standard on identifying discriminative attributes \cite{semeval2018task10}. The dataset consists of 2340 triples (pivot, comparison and attribute) terms, classified as either discriminative (if the attribute is a discriminative feature of pivot) or not (otherwise). The evaluation aimed at answering the following questions:

\noindent \textbf{Q1:} Can the combination of the Definition-Based Model (DBM), the Visual Feature Model (VFM), and the Commonsense Knowledge Graphs (CKG) support the task of discriminative attribute identification?

\noindent \textbf{Q2:} What is the contribution of each component for each type of attribute category?

In the experiment, DBM was instantiated using WordNet 3.0 (only glosses structured into its semantic roles), VFM used VisualGenome 1.4 and CKG used ConceptNet 5. Table \ref{table:categorizedPerf} depicts the outcome of the evaluation of the combined model.

In order to answer Q1, we compared our model to a set of reference systems in the task of discriminative attribute identification (see table \ref{table:SemEvalResults}. While EDAM performs lower than the state-of-the-art (F1-Score 0.76 vs 0.69), from existing approaches EDAM provides the only explainable model. Compared to EDAM, all the top performing models used distributional methods derived from large-scale corpora, while EDAM uses the combination of definition, visual features and commonsense structured KBs.

\begin{table*}
	\scriptsize
	
    \centering
    \begin{center}
        \begin{tabularx}{\textwidth}{ |p{0.2\textwidth}|p{0.5\textwidth}|p{0.12 \textwidth}|p{0.07\textwidth}| }
        \hline
            \textbf{Model} & \textbf{Approach} & \textbf{Explainability} & \textbf{F1 Score} \\
            \hline
            \cite{sunnynlp}&SVM with GloVe&None&0.76\\
            \hline
            \cite{luminoso}&SVM with ConceptNet, Wikipedia articles and WordNet synonyms&None&0.74\\
            \hline
            \cite{ntunlp}&MLP combining information from various DSMs, PMI, and ConceptNet&None&0.73\\
            \hline
            \cite{bomji}&Gradient boosting with co-occurrence count features and \textit{JoBimText} features&None&0.73\\
            \hline
            \cite{uwb}&LexVec, word co-occurrence, and ConceptNet data combined using maximum entropy classifier&None& 0.72\\
            \hline
            \textbf{Proposed Model (EDAM)} & Composes explicit vector spaces from WordNet Definitions, ConceptNet and Visual Genome  & Fully Explainable & 0.69\\
            \hline
            \cite{meaningspace}&Word2Vec cosine similarities of WordNet glosses& Transp. (No expl.) & 0.69\\
            \hline
            \cite{elirfupv}&Use of Wikipedia and ConceptNet& Transp. (No expl.) &0.69\\
            \hline
            \cite{ghh}&Google 5 grams and Word2Vec embeddings as features for feedforward neural network&None&0.67\\
            \hline
            \cite{ecnu}&Ensemble ML model with WordNet, PMI scores, Word2Vec, and GloVe embeddings&None&0.67\\
            \hline
            \cite{discriminator}&A combination of GloVe and Paragram embeddings&None&0.67\\
            \hline
            \cite{umd}&SVM with GloVe embeddings&None&0.67\\
            \hline
            \cite{amritanlp}&CNN with GloVe embeddings&None&0.66\\
            \hline
            \cite{igevorse}&Similarity calculations using a combination of DSMs&None&0.65\\
            \hline
            \cite{thungn}&Word2Vec, GloVe, and FastText embeddings as features for MLP-CNN&None&0.63\\
            \hline
     		\cite{citiusnlp}&Dependency parsing and co-occurrence analysis& Transp. (No expl.) &0.63\\
            \hline
            \cite{alb}&SVM with word co-occurrence&None&0.63\\
            \hline
            \cite{abdn}&CBOW and skip-gram with WordNet definitions& Transp. (No expl.)& 0.62\\
            \hline
            \cite{unbnlp}&Word2Vec, WordNet, and co-occurrence scores&Transp. (No expl.) &0.61\\
            \hline
            \cite{unam}&Convex Cone Method applied to GloVe embeddings&None&0.60\\
     		\hline
		\end{tabularx}
    \end{center}
    \caption{Performance of EDAM in contrast to baselines for discriminative attribute identification. \textit{Transp.} means transparent covering different dimensions of interpretability \cite{VivianInterp} (but without an explanation).}
    \label{table:SemEvalResults}
\end{table*}

\normalsize

In order to answer Q2, we evaluated the performance of each component of the model for the six dual-categories. The goal is to provide a quantitative basis to understand the contribution of each component in addressing the task. 

In order to perform the evaluation, we selected a stratified random subset of 230 triples and manually classified each triple with the dual categories. A triple can contain one or more categories associated with it. The annotation was performed by two independent annotators, which reached an inter-annotator agreement of 81\%. For the final annotated dataset, triples which were not consensual were eliminated (i.e. the final dataset has a 100\% inter-annotator agreement). The F1-score of the combined EDAM model and of each component for each category are shown in Table \ref{table:categorizedPerf}. 

\begin{table*}[!htb]
    \centering
        \begin{tabularx}{\textwidth}{ |X|c|c|c|c|c|c| }
            \hline
            \textbf{Model} & \textbf{Sensory} & \textbf{Logical} & \textbf{Relative} & \textbf{Absolute} & \textbf{Essential} & \textbf{Incidental} \\
            \hline
            DBM & \textbf{0.49} & 0.33 & \textbf{0.40} & 0.41 & 0.29 & \textbf{0.46} \\
            CKG & 0.28 & \textbf{0.50} & 0.20 & \textbf{0.44} & \textbf{0.58} & 0.31 \\
            VFM & 0.13 & 0.11 & 0.05 & 0.14 & 0.10 & 0.13 \\
            EDAM(DBM+CKG+VFM) & 0.62 & 0.63 & 0.50 & 0.65 & 0.68 & 0.60 \\
			\hhline{|=|=|=|=|=|=|=|}
            EDAM gain     & 27\% & 26\% & 25\% & 48\% & 17\% & 30\% \\
            \hline
            \end{tabularx}
    \caption{Performance comparison (recall) against a random sample of categorized triples. The last row shows the relative gain of the combined model over the best performing individual model.}
    \label{table:categorizedPerf}
\end{table*}

The list below summarizes the best component performance for each category:

\begin{itemize}
\item \textbf{Sensory:} DBM significantly outperforms the CKG and VFM model (75\% over the second best). 
\item \textbf{Logical:} CKG significantly outperforms the other components (outperforms the second best component by 52\%). 
\item \textbf{Relative:} DBM significantly outperforms the other components (100\% over the second best).
\item \textbf{Absolute:} CKG outperforms the second best component by 7\%.   
\item \textbf{Essential:} CKG significantly outperforms DBM and VFM (100\% over the second best).
\item \textbf{Incidental:} DBM outperforms VFM by 154\% and CKG by 48\%.
\end{itemize}

The analysis shows the complementary nature of the models, where \textit{DBM} contributes more for the identification of the sensory and incidental categories, while \textit{CKG} contributes to the logical and essential categories. Additionally, \textit{VFM} provides contributions across all categories and has smaller overlaps with other models. It is important to emphasize that the models are complementary at each dimension: the combined model (EDAM) significantly outperforms the best individual models at each dimension (as it can be observed at the gain row on Table \ref{table:categorizedPerf}).

In order to observe the proportion on which each model contributes (and conversely, the amount of redundancy for each model), we analyze for each individual model, the pairwise overlap, in terms of true positives and false positives (Table \ref{table:totalOverl}). The average redundancy between the individual models is low (average 11\%), showing that all models contribute significantly to the performance of the combined model. 

\begin{table*}[!htb]
    \centering
    \begin{center}
        \begin{tabularx}{\textwidth}{|X|c|c|c|c|c|}
            \hline
            \textbf{Positives} & \textbf{DBM} $\land$ \textbf{CKG} & \textbf{DBM} $\land$ \textbf{VFM} & \textbf{CKG} $\land$ \textbf{VFM} & \textbf{DBM} $\land$ \textbf{CKG} $\land$ \textbf{VFM} & \textbf{Avg.}\\
            \hline
            True & 0.23 & 0.07 & 0.08 & 0.06  & 0.11 \\
            False & 0.05 & 0.01 & 0.02 & 0.01  & 0.02 \\
            \hline
        \end{tabularx}
    \end{center}
    \caption{Overlap between the components relative to the combined model.}
    \label{table:totalOverl}
\end{table*}

Table \ref{table:categorizedOverl} further breaks down the true positives for each dimension, where we can analyse the pairwise contribution of each model for each attribute category. \textit{The analysis shows that the overall redundancy between the three models is low for all categories}. Excluding the relative category (in which all models perform poorly), there is little variance in the overlap between the pairwise models (where the intersection between DBM and VFM is the largest for each category). 

\begin{table*}[!htb]
    \centering
    \begin{center}
        \begin{tabularx}{\textwidth}{ |X|c|c|c|c|c| }
            \hline
            \textbf{Category} & \textbf{DBM} $\land$ \textbf{CKG} & \textbf{DBM} $\land$ \textbf{VFM} & \textbf{CKG} $\land$ \textbf{VFM} & \textbf{DBM} $\land$ \textbf{CKG} $\land$ \textbf{VFM} & \textbf{Average}\\
            \hline
            Sensory & 0.31 & 0.10 & 0.14 & 0.10 & 0.16\\
           	Logical & 0.35 & 0.09 & 0.15 & 0.09 & 0.17\\
            Relative & 0.20 & 0.10 & 0.10 & 0.10 & 0.13 \\
            Absolute & 0.36 & 0.09 & 0.15 & 0.09 & 0.15\\
            Essential & 0.33 & 0.05 & 0.10 & 0.05 & 0.13\\
            Incidental & 0.33 & 0.12 & 0.17 & 0.12 & 0.19\\
            \hhline{|=|=|=|=|=|=|}
            Avg. & 0.31 & 0.09 & 0.14 & 0.09 &  \\
			\hline
		\end{tabularx}
    \end{center}
    \caption{Categorical model overlap breakdown, relative to all true positives identified by the combined model.}
    \label{table:categorizedOverl}
\end{table*}

The quantitative analysis shows that the main limitation of the model is in the identification of \textit{relative attributes}. While DBMs are able to correctly identify a small set of triples with relative features such as \texttt{(skyscraper, apartment, tall)} as discriminative, interpretation of relative relations requires two types of features which are not targeted by the models, namely: \emph{(i)} the extraction of precise numerical reference points at scale (dealing with variations of dimensional units) and \emph{(ii)} the ability to extrapolate the relations for unobserved lexemes by an explicit mechanism of comparative/transitive reasoning. As a consequence, the current model is able to identify ``skyscraper'' as being taller than an ``apartment'', but fails to identify neither as taller than a ``giraffe''.
	
DBM forms a significant (49\% each) contribution to \emph{sensory} attribute detection. CKG provides a significant (58\%) contribution to \emph{essential}. Additionally, CKG provides a significant (50\%) contribution to \emph{logical} attribute detection.

\subsection{Error Analysis}

\noindent \textbf{Definition Based Models:} False negatives comprise the majority (83\%) of all model errors. The cause is that the pivot's definition does not include the discriminative attribute. False negatives most commonly occurs with incidental features (e.g. \texttt{(potatoes, butter, mashed) - true:false} and \texttt{(nose, throat, perfume) - true:false}). False positives occur when the attribute applies to both pivot and comparison, however the comparison's definition does not include the feature. This is most prevalent with incidental attributes (e.g. \texttt{(banana,onions,peel) - false:true} and \texttt{(torah, bible, read) - false:true}).

\noindent \textbf{Common Sense Knowledge Graphs:} False negatives comprise the majority (80\%) of the model's errors and are mainly caused by incidental attributes (e.g. \texttt{(trays,employee,wooden) - true:false}) and relative attributes (e.g. \texttt{(stool,tray,tall) - true:false}).

\noindent \textbf{Visual Feature Models:} Similarly as with DBMs and CKGs, false negatives comprise the majority (94\%) of all false classifications. Most of the errors occur either with logical attributes (e.g. \texttt{(wife, lady, married) - true:false}, since these attributes are not likely to be expressed in a visual corpus or relative attributes (e.g. \texttt{(torah, bible, short) - true:false}. False positives occur due to domain incompleteness. This can be caused by a lack of domain coverage of the Visual Genome dataset (e.g. in \texttt{(cat, lion, whiskers) - false:true} ; \texttt{(meal, supper, food) - false:true}).

\noindent \textbf{Combined EDAM:} \textit{The decrease in the proportion of false negatives as compared to the individual models, illustrates the advantages of the model composition}. False negatives comprise 47\% of model's total errors. False positives, on the contrary, illustrate the limitations of the model, since they occur when at least one of the model components incorrectly classifies a triple as discriminative: e.g. \texttt{(banana, onions, peel) - false:true} false positive of DBM and \texttt{(cat, lion, whiskers) - false:true} false positive of VFM.

\subsection{Qualitative Analysis}

In addition to correctly labeling common discriminative triples such as
\texttt{(soup, meal, liquid)} and \texttt{(walnut, spinach, brown)} the technical specificity of definitions present at dictionaries supports the identification of discriminative triples such as \texttt{(brandy, whiskey, wine)}. Other noteworthy examples of discriminative attribute detection include labeling triples \texttt{(stomach, bladder, food)} and \texttt{(nightclub, bar, dancing)} as discriminative.

For the DBM, the structure induced by the extractor (e.g. the hypernym hierarchy) can support the transference of discriminative attributes across the taxonomic hierarchy. For the triple \texttt{(planet, moon, body)} using immediate definitions of pivot and comparison incorrectly suggests that the triple \texttt{(planet, moon, body)} is discriminative. However, after expanding using super-type definitions, the model correctly identifies \texttt{body} as a property of both \texttt{planet} and \texttt{moon}.

\section{Conclusion}
\label{sec:con}

This paper described an explicit word vector model targeting the identification of discriminative attributes using the composition of definitions, visual features and commonsense knowledge graphs. The proposed model, which is built from structured representations from different types of data sources is able to achieve a state-of-the-art level F1-score (0.69) while producing, human interpretable explanations. The paper also provided an in-depth comparative quantitative and qualitative analysis on the contributions of different types of data sources for the generation of explicit semantic vector spaces (WordNet glosses, ConceptNet and Visual Genome), demonstrating the complementarity aspect of these resources. Future work will concentrate on extending the model to cope with relative attributes, the inclusion of additional data sources to increase model coverage, such as large-scale definition sets.

\section*{Acknowledgments} The authors would like to thank Viktor Schlegel for the supporting basic machine learning baselines in the beginning of the project (which are not used or reported here). 

\bibliography{bibliography.bib}
\bibliographystyle{acl_natbib}

\appendix

\end{document}